\documentclass[10pt,twocolumn,letterpaper]{article}

\usepackage{iccv}
\usepackage{times}
\usepackage{epsfig}
\usepackage{graphicx}
\usepackage{amsmath}
\usepackage{amssymb}
\usepackage{float}

\usepackage[breaklinks=true,bookmarks=false]{hyperref}

\iccvfinalcopy 

\def\iccvPaperID{****} 

\ificcvfinal\fi

\begin{document}

\title{DiffuEraser: A Diffusion Model for Video Inpainting
\makebox[\textwidth][c]{\small{TECHNICAL REPORT}}
\\[-0.3cm]
}

\author{
\makebox[\textwidth][c]{%
    \begin{tabular}{@{}c@{}}
    Xiaowen Li \and Haolan Xue \and Peiran Ren \and Liefeng Bo\\
    \end{tabular}
  } \\[0.1cm]
\makebox[\textwidth][c]{Tongyi Lab, Alibaba Group}\\
\makebox[\textwidth][c]{\small \texttt{\{lxw262398, haolan.xhl, peiran.rpr, liefeng.bo\}}\texttt{@alibaba-inc.com}}\\
\makebox[\textwidth][c]{\small\url{https://github.com/lixiaowen-xw/DiffuEraser.git}}
}

\maketitle
\ificcvfinal\thispagestyle{empty}\fi

\begin{abstract}
  Recent video inpainting algorithms integrate flow-based pixel propagation with transformer-based generation to leverage optical flow for restoring textures and objects using information from neighboring frames, while completing masked regions through visual Transformers. However, these approaches often encounter blurring and temporal inconsistencies when dealing with large masks, highlighting the need for models with enhanced generative capabilities. Recently, diffusion models have emerged as a prominent technique in image and video generation due to their impressive performance. In this paper, we introduce DiffuEraser, a video inpainting model based on stable diffusion, designed to fill masked regions with greater details and more coherent structures. We incorporate prior information to provide initialization and weak conditioning, which helps mitigate noisy artifacts and suppress hallucinations. Additionally, to improve temporal consistency during long-sequence inference, we expand the temporal receptive fields of both the prior model and DiffuEraser, and further enhance consistency by leveraging the temporal smoothing property of Video Diffusion Models. Experimental results demonstrate that our proposed method outperforms state-of-the-art techniques in both content completeness and temporal consistency while maintaining acceptable efficiency.
\end{abstract}

\section{Introduction}

Video inpainting aims to complete masked regions with content that is both plausible and temporally consistent. Previous video inpainting algorithms primarily rely on two mechanisms:

1) \textbf{Flow-based pixel propagation methods}, which utilize optical flow to restore texture details and objects by leveraging information from adjacent frames; and

2) \textbf{Transformer-based video inpainting methods}, which excel at completing the structural aspects of objects \cite{quan2024deeplearningbasedimagevideo}.

Current mainstream algorithms typically combine these two approaches, consisting of three modules or stages:

1) \textbf{Flow completion},

2) \textbf{Feature propagation}, and

3) \textbf{Content generation}.

This solution categorizes masked pixels into two types:

1) \textbf{Known pixels}, which have appeared in some masked frames and can be propagated to other frames through flow completion and feature propagation modules, ensuring consistency between the completed content and the unmasked regions; and

2) \textbf{Unknown pixels}, which have never appeared in any masked frames and are generated by the content generation module, thereby enhancing the structural integrity of the results.

The state-of-the-art algorithm, \textbf{Propainter} \cite{zhou2023propainter}, exemplifies this approach and comprises three key modules: recurrent flow completion, dual-domain propagation, and mask-guided sparse Transformer. It effectively propagates known pixels across all frames and demonstrates an initial ability to generate unknown pixels. However, when the mask size is large, the generative capability of the Transformer model proves insufficient, leading to significant artifacts, as illustrated in Figure \ref{fig:1}.

Consequently, there is a need for more powerful models with enhanced generative capabilities. The \textbf{Stable Diffusion} model, which has recently gained prominence in the field of image and video generation, presents a promising candidate.

In this work, we first decompose the video inpainting task into three sub-problems and then propose corresponding solutions for each. Specifically, the three key challenges are: the propagation of known pixels, the generation of unknown pixels, and the temporal consistency of the completed content. Our main contributions are summarized as follows:

\begin{enumerate}
    \item \textbf{Video Inpainting Diffusion}: We introduce a motion module for the image inpainting model BrushNet, which is based on diffusion models. The powerful generative capability of diffusion models overcomes the blurring and mosaic artifacts associated with Transformer-based models, thereby completing object structures and generating more detailed content.
    
    \item \textbf{Injected Priors}: We incorporate priors into the diffusion model, enabling easier initialization to mitigate noisy artifacts and serving as a weak condition to suppress the generation of unwanted objects.
    
    \item \textbf{Enhanced Temporal Consistency}: We improve the temporal consistency of long-sequence inference by expanding the temporal receptive fields of both the prior model and the diffusion model. Additionally, we further enhance temporal continuity at the intersections between clips by leveraging the temporal smoothing property of the Video Diffusion Model.
\end{enumerate}

\section{Related Works}

\textbf{Diffusion Models.} The advent of diffusion models \cite{ho2020denoising, sohldickstein2015deepunsupervisedlearningusing, song2021scorebased} has significantly enhanced the quality and creativity of image and video generation. In the realm of image synthesis, diffusion models have driven substantial progress across various tasks, including text-to-image generation \cite{ramesh2022, saharia2022photorealistictexttoimagediffusionmodels}, controllable image generation \cite{mou2023t2i, zhang2023adding}, image editing \cite{brooks2022instructpix2pix, hertz2022prompt, mokady2022null}, personalized image generation \cite{gal2022textual, ruiz2022dreambooth}, and image inpainting \cite{rombach2021highresolution, ju2024brushnet}, among others. Building on these advancements, video diffusion models incorporating additional motion modules have also gained significant traction. Key applications in this domain include text-to-video generation \cite{guo2023animatediff, ge2024preservecorrelationnoiseprior, reuse2023, ho2022imagenvideohighdefinition, ho2022video, singer2022makeavideotexttovideogenerationtextvideo}, controllable video generation \cite{chen2023controlavideo, esser2023structurecontentguidedvideosynthesis, wang2023videocomposercompositionalvideosynthesis, xing2023make}, video editing \cite{liew2023magicedit, molad2023dreamixvideodiffusionmodels, wu2023tune, liu2023videop2p}, and various training-free video synthesis methods \cite{zhang2023controlvideo, qi2023fatezero}.

\textbf{Video Inpainting.} Video inpainting aims to fill masked regions in videos with plausible content while maintaining temporal consistency. Early approaches based on 3D convolution and shifting operations exhibited limited performance. The emergence of methods leveraging optical flow and Transformer architectures has significantly improved the quality of video inpainting. Flow-based pixel propagation methods \cite{Gao-ECCV-FGVC, zhang2022flow, Zhang_2022_CVPR} excel at restoring textures and details by utilizing information from adjacent frames. In contrast, Transformer-based methods \cite{yan2020sttn, Liu_2021_FuseFormer, liCvpr22vInpainting, zhou2023propainter} are adept at completing the structural aspects of objects. Among these, Propainter \cite{zhou2023propainter} stands out as a representative approach, comprising recurrent flow completion, dual-domain propagation, and a mask-guided sparse Transformer. Propainter effectively propagates known pixels across all frames and demonstrates an initial ability to generate unknown pixels. However, its generative capacity is limited when dealing with large masks, leading to noticeable artifacts.

\begin{figure*}[ht]
\begin{center}
\includegraphics[width=0.8\linewidth]{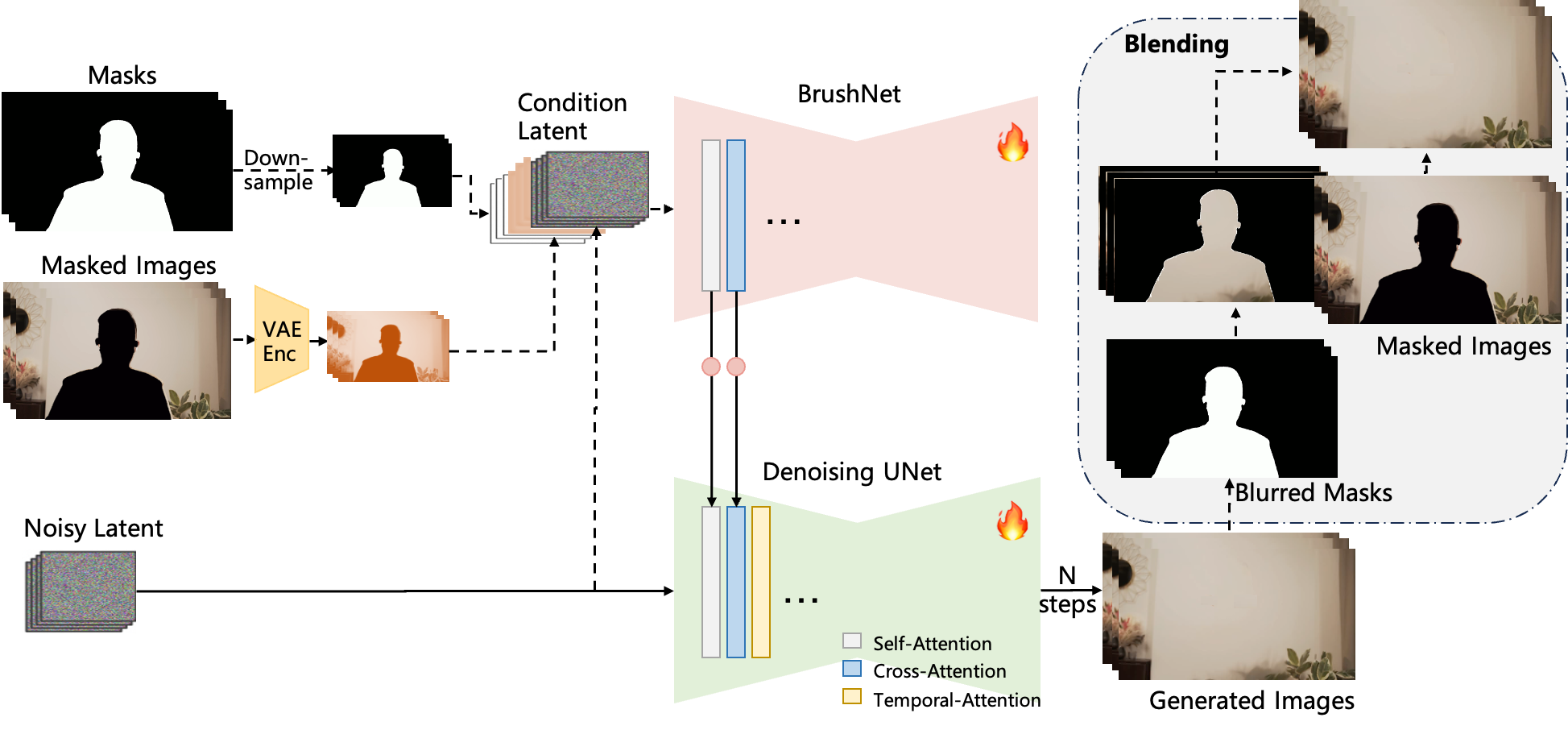}
\end{center}
   \caption{Overview of the proposed video inpainting model \textbf{DiffuEraser}, based on stable diffusion. The main denoising UNet performs the denoising process to generate the final output. The BrushNet branch extracts features from masked images, which are added to the main denoising UNet layer by layer after a zero convolution block. Temporal attention is incorporated after self-attention and cross-attention to improve temporal consistency.}
\label{fig:2}
\end{figure*}

With the rising popularity of diffusion models, diffusion-based video inpainting methods have begun to emerge \cite{lee2024videodiffusionmodelsstrong, wu2024lgvi, Shi_2024_CVPR, gu2024advanced, zhang2023avid, Zi2024CoCoCo}. These approaches leverage the powerful generative capabilities of diffusion models to enhance both the detail and structural integrity of the inpainted regions, addressing some of the limitations observed in Transformer-based methods. BIVDiff\cite{Shi_2024_CVPR} is a training-free framework via bridging image and video diffusion models. AVID\cite{zhang2023avid} and CoCoCo\cite{Zi2024CoCoCo} improved text-guided video inpainting by integrating motion module to Text-to-Image(T2I) model. \cite{wu2024lgvi} proposes language-driven video inpainting via Multimodal Large Language Models, which uses natural language instructions to guide the inpainting process. Nevertheless, they always suffer from the inherent hallucinations of diffusion models. FloED\cite{gu2024advanced} with less hallucination proposes a dedicated dual-branch architecture that incorporates motion guidance with a multi-scale flow adapter to enhance temporal consistency, focusing on object removal and background restoration. FFF-VDI\cite{lee2024videodiffusionmodelsstrong} propagates the noise latent information of future frames to fill the masked area of the first frame's noise latent code, improving temporal consistency and suppressing hallucination effects. However, these methods do not effectively address the temporal consistency and stability needed for long-sequence inference and there is still room for improvement in detail and structural integrity. In contrast, DiffuEraser can generate temporally consistent results with enhanced detail and a more complete structure for long-sequence inference, all without requiring a text prompt.

\section{Methodology}

\subsection{Network Overview}

Our network architecture is inspired by AnimateDiff \cite{guo2023animatediff}, integrating a motion module into the image inpainting model. For the image inpainting component, we select BrushNet \cite{ju2024brushnet}, which enhances the main denoising UNet by adding an additional branch to extract features from masked images. An overview of our proposed model, \textbf{DiffuEraser}, is depicted in Figure \ref{fig:2}. The architecture comprises the primary denoising UNet and an auxiliary BrushNet. The BrushNet branch receives a conditional latent input composed of masked images, masks, and noisy latents, with dimensions \([n, f, h/4, w/4, 9]\). Features extracted by BrushNet are integrated into the denoising UNet layer by layer after a zero convolution block. The denoising UNet processes noisy latents with dimensions \([n, f, h/4, w/4, 4]\). To enhance temporal consistency, temporal attention mechanisms are incorporated following both self-attention and cross-attention layers. After denoising, the generated images are blended with the input masked images using blurred masks.

We define the video inpainting problem by decomposing it into three sub-problems: propagation of known pixels (pixels that have appeared in some masked frames), generation of unknown pixels (pixels that have never appeared in any masked frames), and maintaining temporal consistency of the completed content. Specifically:

\begin{enumerate}
    \item \textbf{Propagation of Known Pixels:} 
    The motion module inherently supports temporal propagation, allowing the restoration of texture details and objects in the current frame using information from adjacent frames. Additionally, we leverage the enhanced propagation capabilities of the prior model, which offers a longer propagation range and a more sophisticated propagation mechanism. Specifically, we apply DDIM inversion on the inpainting results from the prior model and incorporate them into the noisy latent. See Section \ref{section:prior} for details. We utilize Propainter as our prior model. Beyond supporting the propagation of known pixels, the injected prior facilitates easier initialization for DiffuEraser, enabling the generation of meaningful completed content and suppressing noisy artifacts and visual hallucinations commonly associated with diffusion models.
    
    \item \textbf{Generation of Unknown Pixels:} 
    Utilizing the robust generative capabilities of the stable diffusion model, our approach can generate plausible content with more details and textures for unknown pixels.
    
    \item \textbf{Temporal Consistency of Completed Content:} 
    While the motion module ensures temporal consistency within individual inferences (each handling a clip of 22 frames in our setting), discrepancies arise at the boundaries between clips during long-sequence processing. To address this, we expand the temporal receptive field of the model. This is achieved by performing pre-inference, where video frames are sampled at an optimal rate and processed collectively as a single clip. This enables the model to "see" frames from a broader temporal context. Subsequently, the insights gained from pre-inference are used to guide the frame-by-frame inference, incorporating information from distant frames and thereby enhancing the overall temporal continuity. See Section \ref{section:temporal-consistency} for details.
\end{enumerate}

As demonstrated in other studies, the generative capability of stable diffusion models and the temporal consistency provided by motion modules are well-established. In this paper, we focus on illustrating the advantages of incorporating priors and optimizing temporal consistency across clips during long-sequence inference.

\subsection{Incorporation of Priors}
\label{section:prior}

\begin{figure}[ht]
\begin{center}
\includegraphics[width=0.9\linewidth]{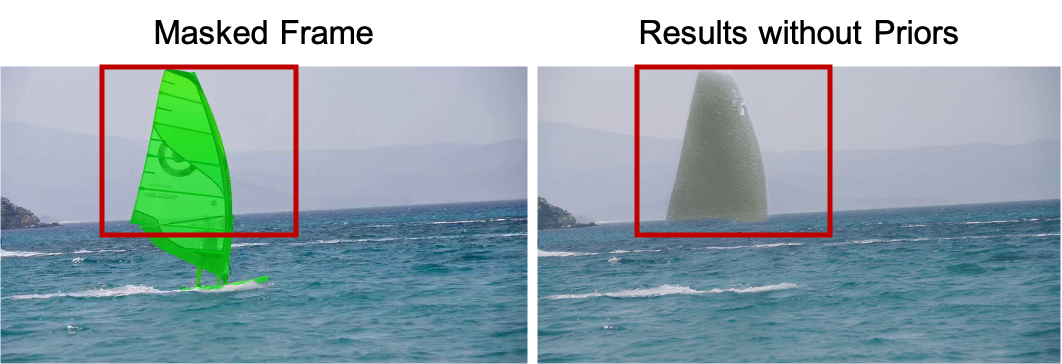}
\end{center}
   \caption{Example of noisy artifacts generated by the model. The masked region above the sea level is not completed correctly and resembles random noise.}
\label{fig:3}
\end{figure}

As illustrated in Figure \ref{fig:3}, our model occasionally generates meaningless noisy artifacts within masked regions. For instance, the masked area above the sea level may appear as random noise instead of coherent content.

\begin{figure}[ht]
\begin{center}
\includegraphics[width=1.0\linewidth]{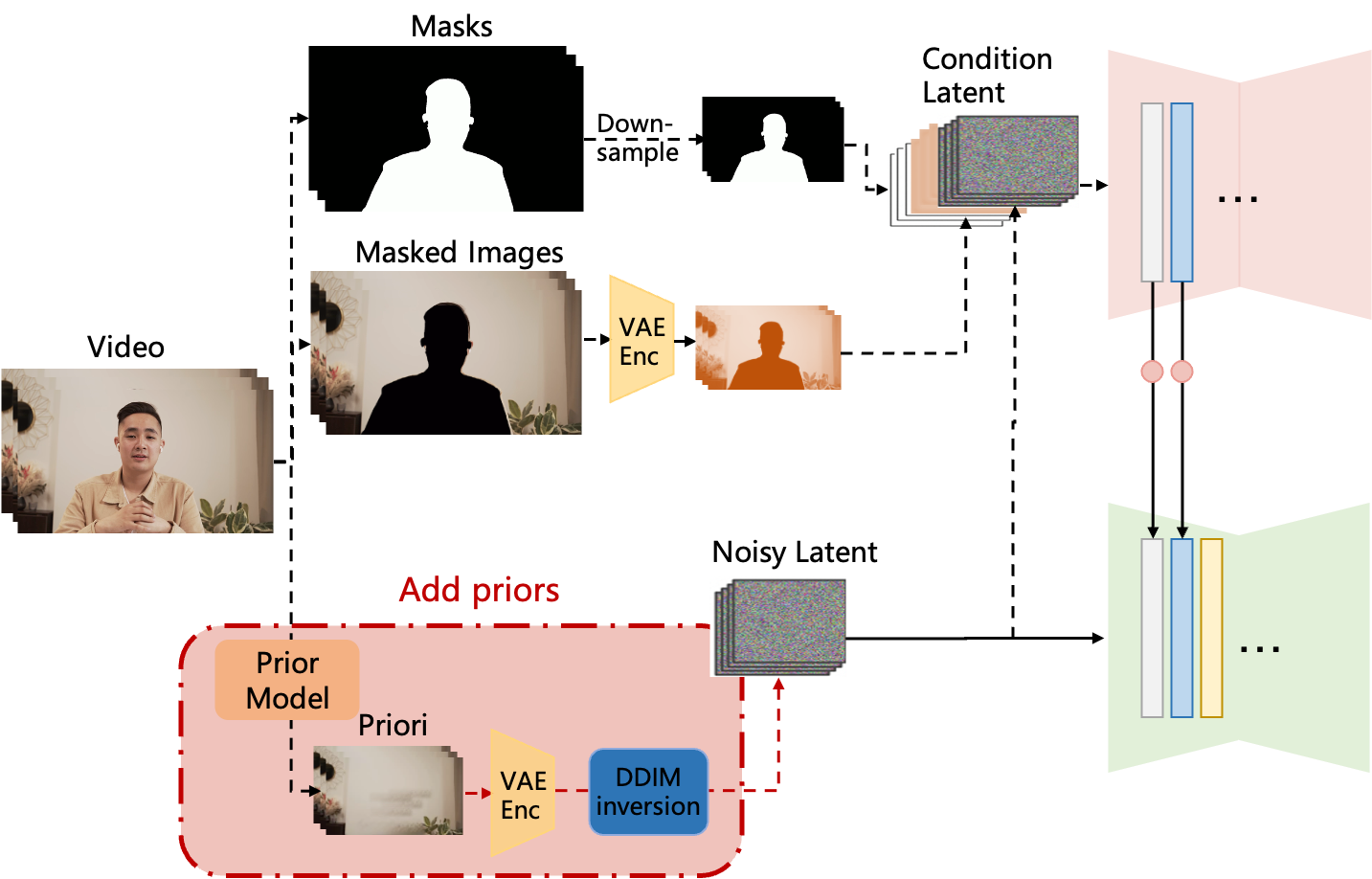}
\end{center}
   \caption{Incorporation of priors. We introduce priors during inference by performing DDIM inversion on the outputs of the prior model and adding them to the noisy latent.}
\label{fig:4}
\end{figure}

To address these artifacts, we enhance the noisy latent—an integral part of the model's input. Inspired by DDIM Inversion \cite{song2020denoising}, we introduce priors during inference. Specifically, we perform DDIM Inversion on the outputs of a chosen lightweight inpainting model and incorporate the inverted results into the noisy latent, as depicted in Figure \ref{fig:4}. The prior provides initialization information that enables the model to generate meaningful and stable completed content, effectively eliminating the noisy artifacts shown in Figure \ref{fig:3}. Additionally, the prior acts as a weak condition to suppress the generation of unwanted objects, mitigating visual hallucinations often encountered in diffusion models.

The selection of the prior model significantly impacts the final results. After experimental comparisons, we selected Propainter as our prior model. Notably, any blur and mosaic artifacts present in the prior do not adversely affect our model's outputs; instead, they are refined and eliminated, resulting in inpainted regions with richer textures and greater detail.

\begin{figure}[ht]
\begin{center}
\includegraphics[width=0.9\linewidth]{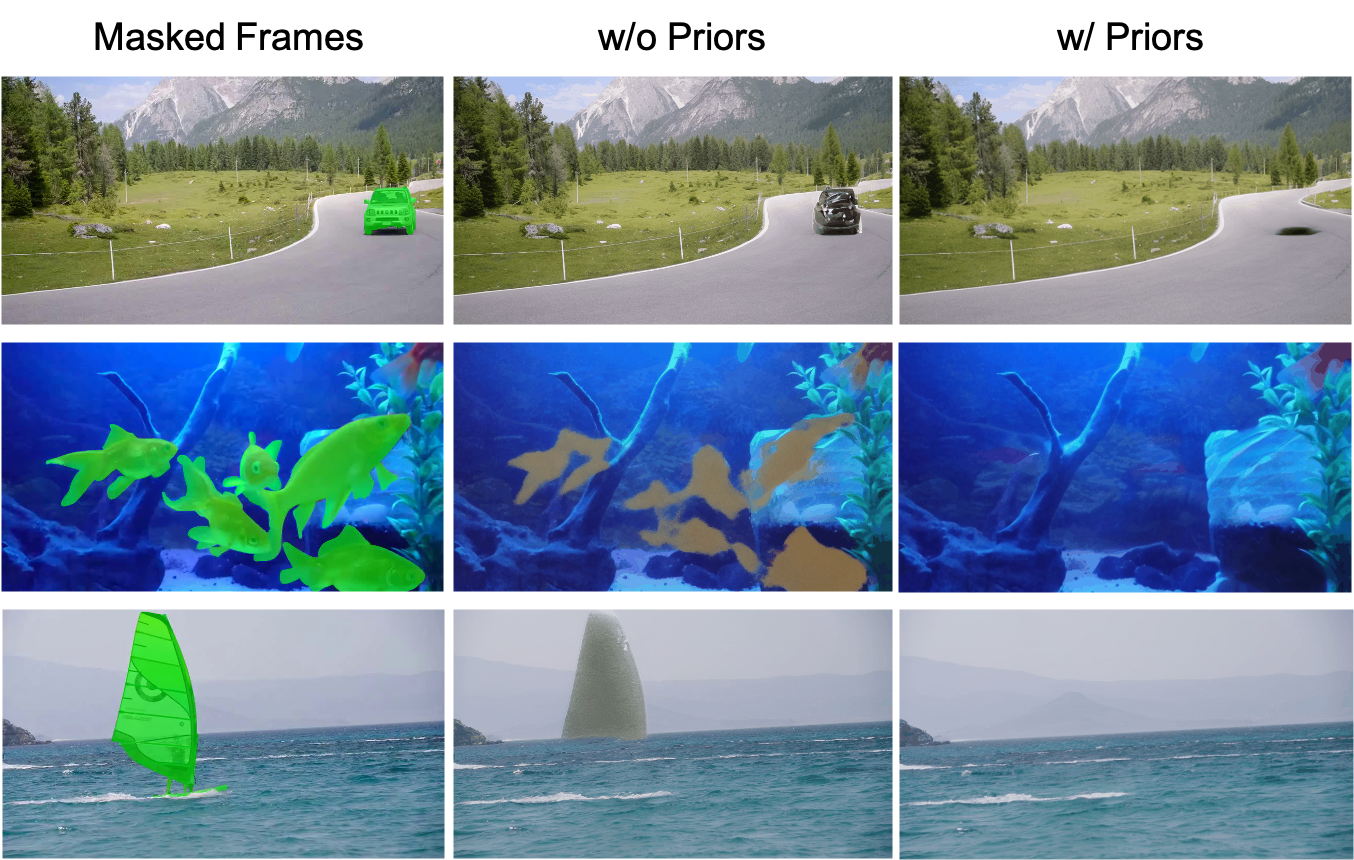}
\end{center}
   \caption{Comparison of inpainting results before and after incorporating priors.}
\label{fig:5}
\end{figure}

Figure \ref{fig:5} compares the results before and after incorporating priors, demonstrating that the introduction of priors effectively suppresses noisy artifacts and the emergence of unwanted objects, thereby significantly enhancing the accuracy and stability of the inpainting results.

\begin{figure*}[ht]
\begin{center}
\includegraphics[width=1.0\linewidth]{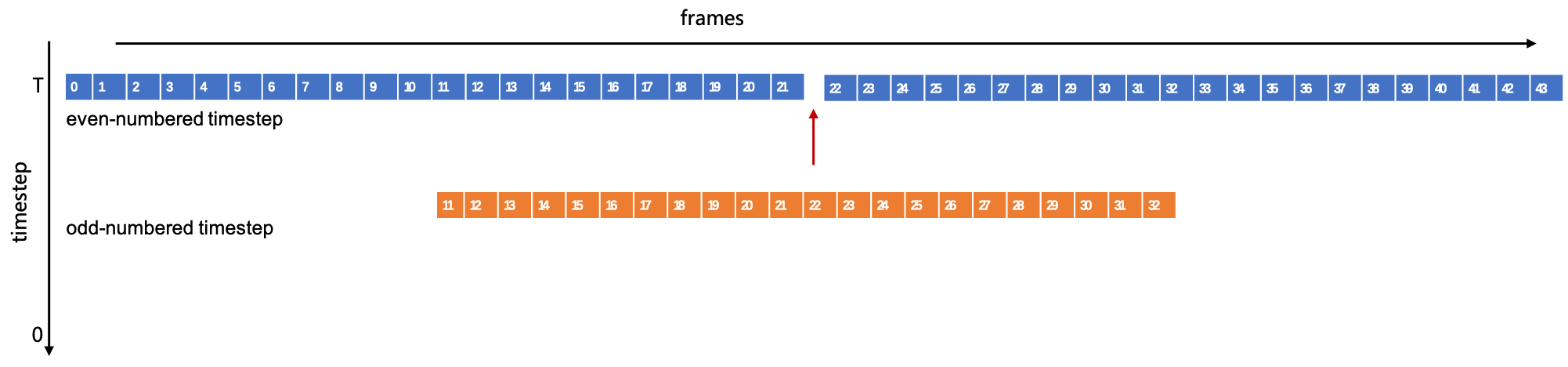}
\end{center}
   \caption{Utilizing the temporal smoothing property of the Video Diffusion Model (VDM) to enhance consistency at the intersections of clips.}
\label{fig:6}
\end{figure*}

\subsection{Optimizing Temporal Consistency for Long-Sequence Inference}
\label{section:temporal-consistency}

\begin{figure}[ht]
\begin{center}
\includegraphics[width=1.0\linewidth]{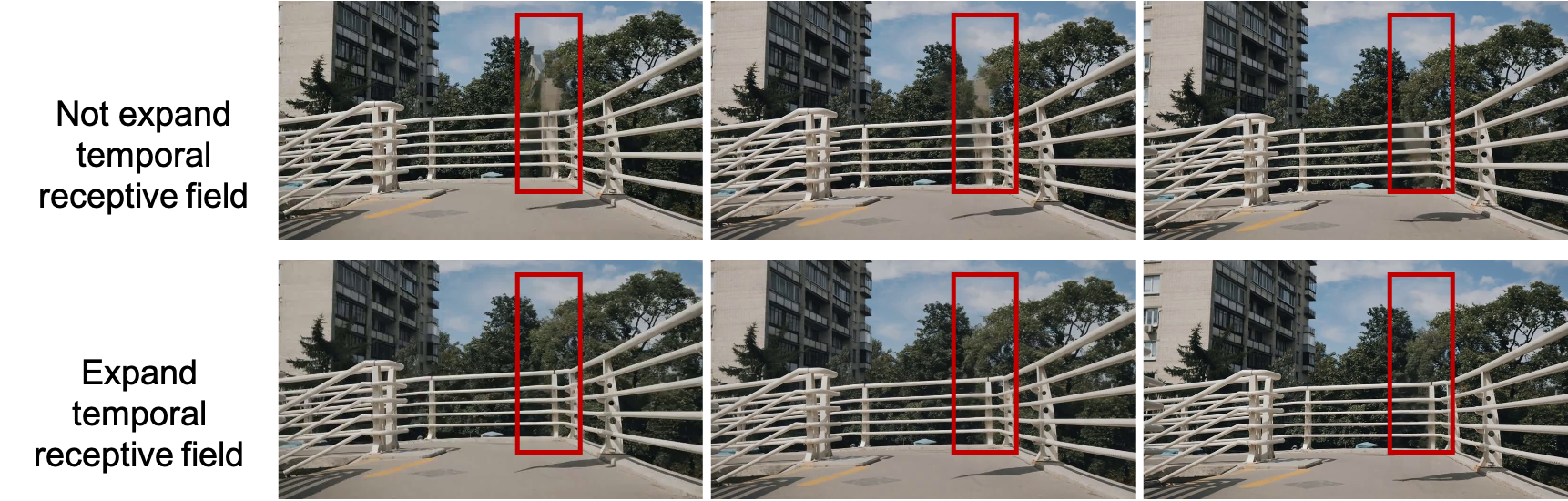}
\end{center}
   \caption{Temporal consistency optimization for long-sequence inference.}
\label{fig:7}
\end{figure}

While the motion module maintains good temporal consistency within individual clips(for example, 22 frames), noticeable discrepancies emerge at the boundaries between consecutive clips during long-sequence inference, as shown in Figure \ref{fig:7}. To ensure seamless temporal consistency across the entire video, we implement the following optimizations.

\subsubsection{Leveraging the Temporal Smoothing Property of the Video Diffusion Model (VDM)}

The absence of specific temporal conditioning leads to significant changes in completed content between clips, a problem that cannot be resolved by merely overlapping neighboring clips. Inspired by the concept of interpolating between timesteps to obtain intermediate results \cite{gu2024advanced}, we adopt a staggered denoising approach along sequential timesteps. This method utilizes the inherent temporal smoothing property of VDM to enhance consistency between clips.

During inference, even-numbered timesteps remain inferred from the starting position of the clip, while odd-numbered timesteps are inferred from the midpoint of the clip, Figure \ref{fig:6}. This staggered denoising leverages VDM's temporal smoothing property to blend frames at clip intersections smoothly. The underlying rationale is that, despite identical latent inputs, the denoising results for overlapped frames from adjacent clips differ due to VDM's temporal smoothing property, which adjusts overlapped frames to be temporally consistent with the starting frame. By applying this smoothing property at clip intersections, we achieve more seamless transitions.

When processing long videos divided into multiple clips, preliminary optimizations lead to multiple adjustments at clip intersections. After optimization, these transitions are smoothed into a single gradual change from the first to the last frame of the entire video. However, complete consistency across the entire video remains unattainable due to inherent inconsistencies between the first and last frames.

\subsubsection{Expanding the Temporal Receptive Field}

A single inference pass can process only a limited number of frames(for instance, 22 frames in our setting), which restricts the temporal receptive field and prevents the propagation of known pixels from distant frames. Additionally, information sharing between different clips is constrained, resulting in inconsistencies in detailed content despite similar semantics across clips. This leads to frequent and noticeable changes during long-sequence inference, as illustrated in Figure \ref{fig:7}. To mitigate this, we expand the temporal receptive field of the inference process through the following two strategies.

\textbf{1. Enhancing Priors for Comprehensive Pixel Propagation}

\begin{figure}[ht]
\begin{center}
    \includegraphics[width=1.0\linewidth]{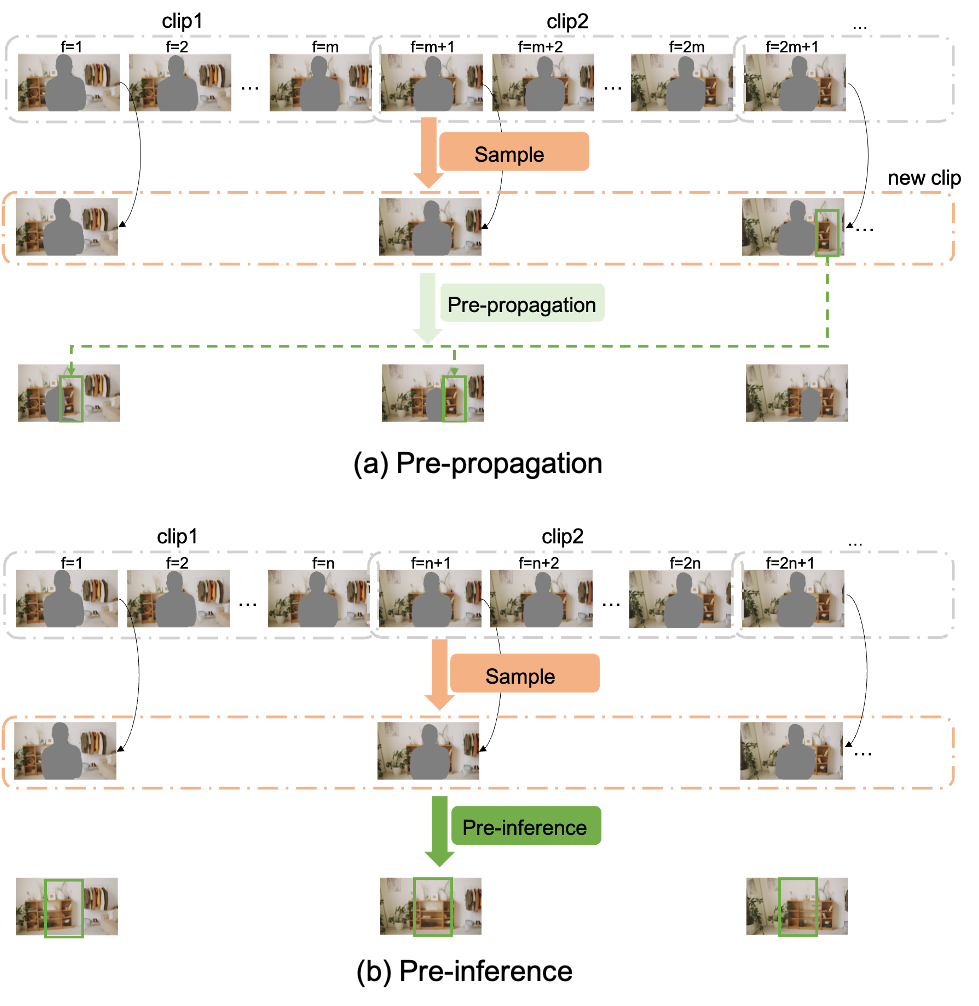}
    \vspace{-20pt}
\end{center}
   \caption{Perform pre-propagation or pre-inference to expand the temporal receptive field of model.}
\label{fig:89a}
\end{figure}

Using Propainter as an example, we first sample the input video frames and perform pre-propagation to extend known pixels across the entire time domain, surpassing the temporal limitations of a single propagation pass (which typically handles dozens of frames), as shown in Figure \ref{fig:89a}(a). Full propagation of known pixels ensures that the completed content remains consistent with the unmasked regions, thereby stabilizing the results.

\begin{figure}[ht]
\begin{center}
    \includegraphics[width=1.0\linewidth]{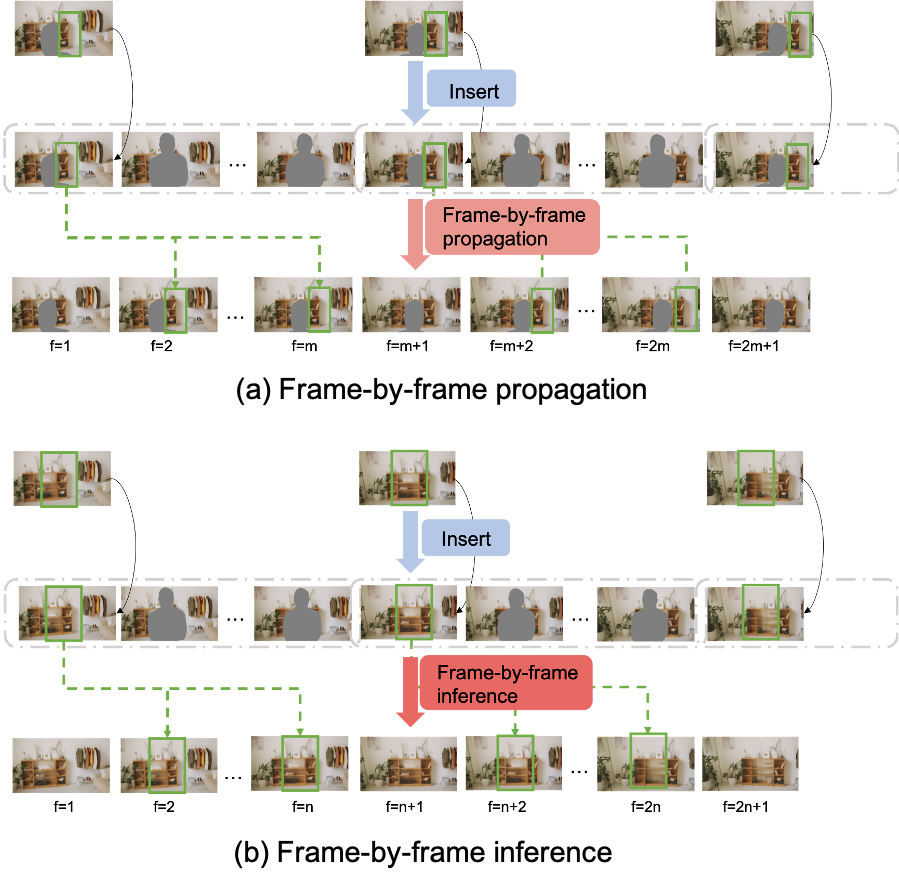}
    \vspace{-20pt}
\end{center}
   \caption{The temporal consistency obtained from pre-propagation or pre-inference is maintained throughout all remaining frames.}
\label{fig:89b}
\end{figure}

Subsequently, the inpainting results of the sampled frames guide frame-by-frame propagation, allowing the information obtained from pre-propagation to be integrated into every frame, as depicted in Figure \ref{fig:89b}(a).

This optimization enables Propainter to utilize information from distant frames more effectively, ensuring that known pixels are stably propagated across the entire time domain. Consequently, the prior provided to DiffuEraser is more accurate and stable. Nonetheless, DiffuEraser's limited temporal receptive field still results in significant changes at clip intersections.

\textbf{2. Expanding the Temporal Receptive Field of DiffuEraser for consistent generation
of unknown pixels}

To further enhance temporal consistency, we also expand the temporal receptive field of DiffuEraser. Similar to the prior optimization, we introduce a pre-inference step where video frames are sampled and processed as a single inference pass, thereby broadening the temporal context and ensuring consistent content generation across the entire video, as shown in Figure \ref{fig:89a}(b).

Following pre-inference, the results guide frame-by-frame inference, ensuring that the content consistency established during pre-inference is maintained throughout all remaining frames, as illustrated in Figure \ref{fig:89b}(b).

The core principle behind these optimizations—both for priors and DiffuEraser—is to extend the temporal receptive field to encompass the entire video duration, rather than being confined to individual clips. The optimization of prior ensures comprehensive propagation of known pixels, maintaining result correctness, while the optimization of DiffuEraser focuses on the consistent generation of unknown pixels, ensuring overall stability. Together, these enhancements effectively resolve the temporal consistency issues inherent in long-sequence inference, as demonstrated in Figure \ref{fig:7}.


\section{Experiments}

\textbf{Datasets.} We utilized the Panda-70M dataset \cite{chen2024panda70m}, splitting videos at scene cuts and filtering them based on matching scores to obtain 3,183,727 short video clips paired with captions. During training, we generated mask sequences with random rates, directions, and shapes to simulate video inpainting and object removal tasks.

\textbf{Training Details and Metrics.} We employed a two-stage training strategy with a resolution of 512. In the first stage, we trained the BrushNet and the main denoising UNet without the motion module to enhance content generation capabilities. In the second stage, we trained the motion module of the main denoising UNet to improve temporal consistency. The fist stage is trained on 4 NVIDIA A100 GPUs for 100,000 steps with a batch size of 16, and the second stage is trained on 8 NVIDIA A100 GPUs for 80,000 steps with 22-frame video sequences and a batch size of 1. Both models were optimized using the L2 loss function and a learning rate of 1e-5.

\textbf{Efficiency.} Leveraging Phased Consistency Models (PCM) \cite{wang2024phased}, our model can generate samples in only two steps, significantly improving inference efficiency. For instance, processing a 10-second video at 540p and 25 FPS using Nvidia GPU L20 requires about 200 seconds.

\begin{figure*}[ht]
\begin{center}
    \includegraphics[width=0.95\linewidth]{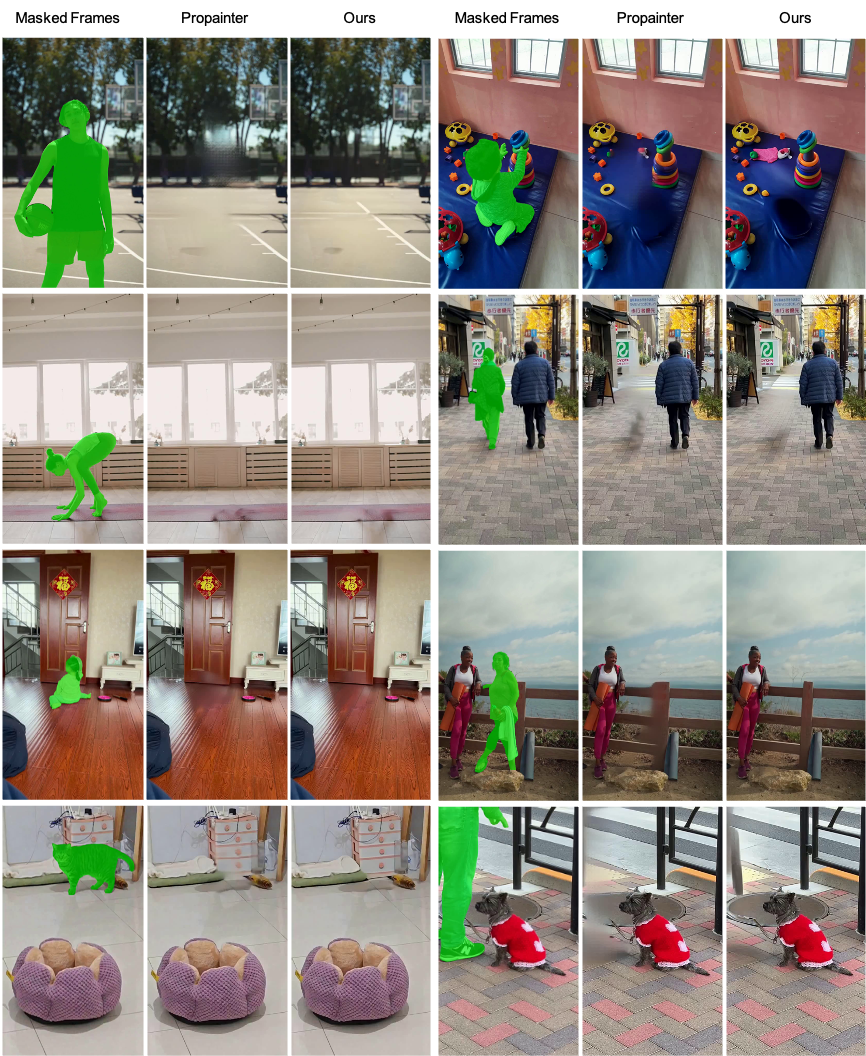}
\end{center}
   \caption{Texture quality comparison between \textbf{DiffuEraser} and Propainter.}
   \vspace{10pt}
\label{fig:11}
\end{figure*}

\begin{figure*}[ht]
\begin{center}
    \includegraphics[width=0.95\linewidth]{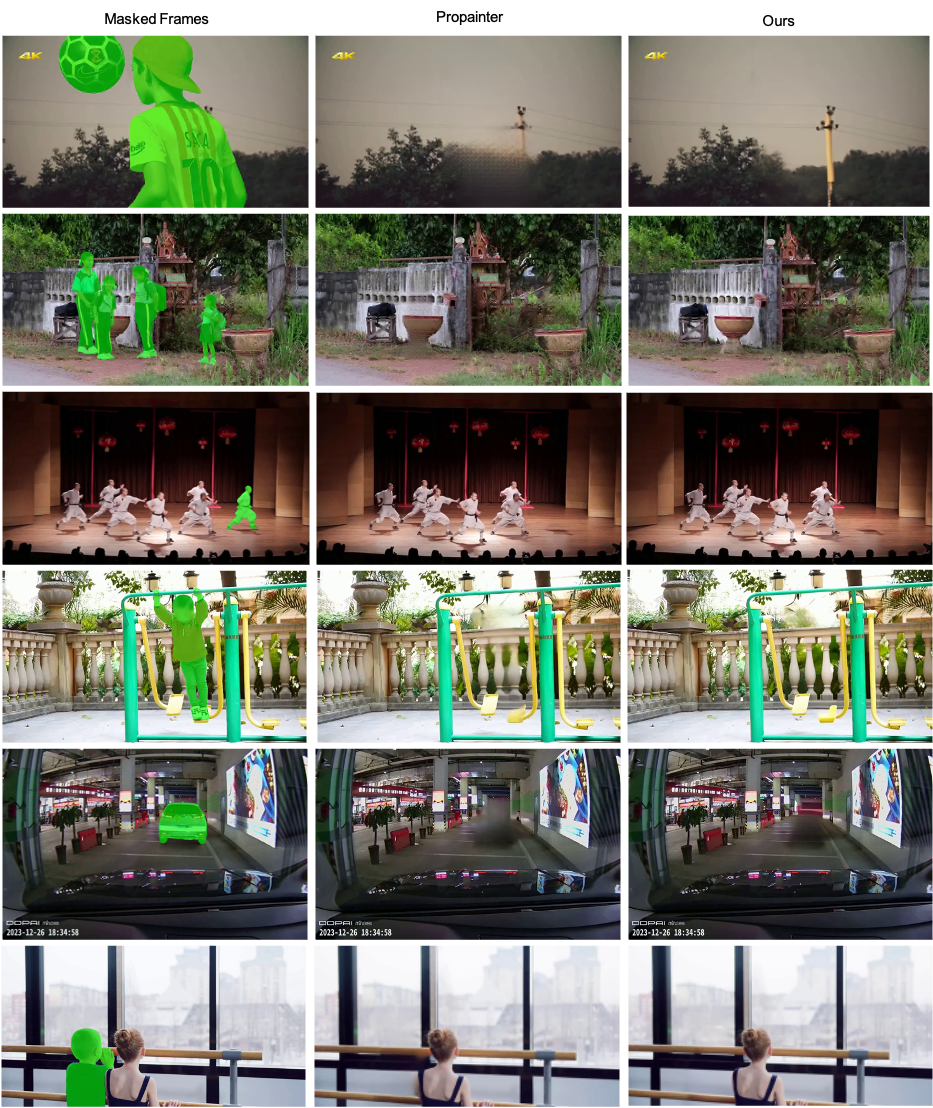}
\end{center}
   \caption{Texture quality comparison between \textbf{DiffuEraser} and Propainter.}
   \vspace{30pt}
\label{fig:10}
\end{figure*}

\begin{figure*}[ht]
\begin{center}
    \vspace{-4pt}
    \includegraphics[width=0.65\linewidth]{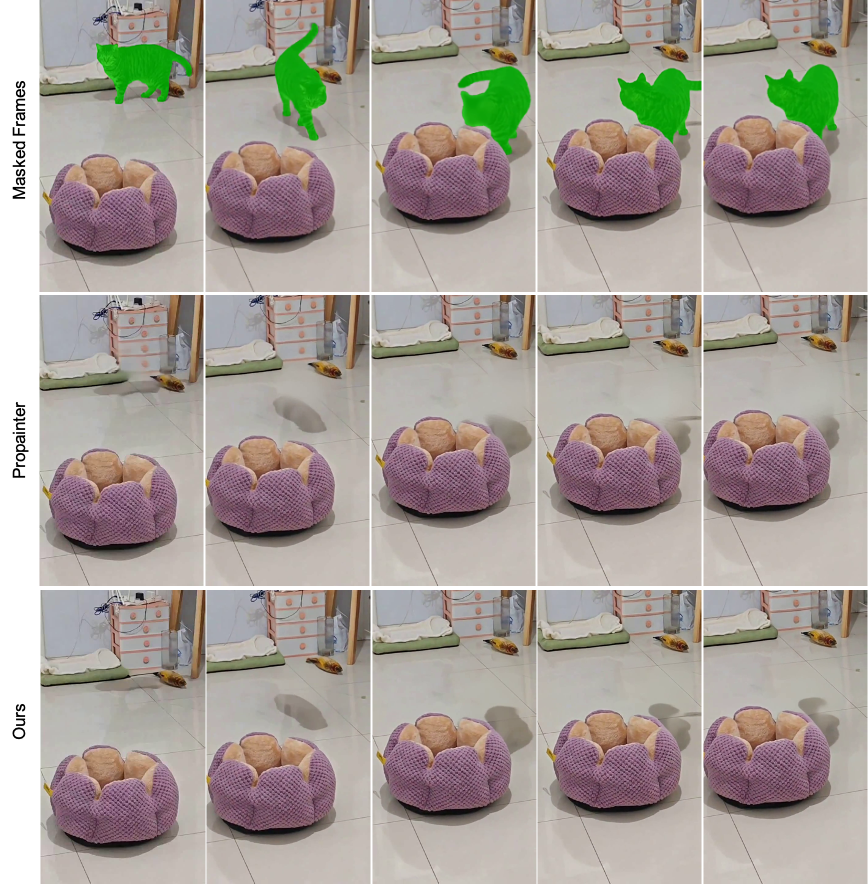}
    \vspace{-8pt}
\end{center}
   \caption{Temporal consistency comparison between \textbf{DiffuEraser} and Propainter.}
\label{fig:13}
\end{figure*}

\begin{figure*}[ht]
\begin{center}
    \vspace{-5pt}
    \includegraphics[width=0.9\linewidth]{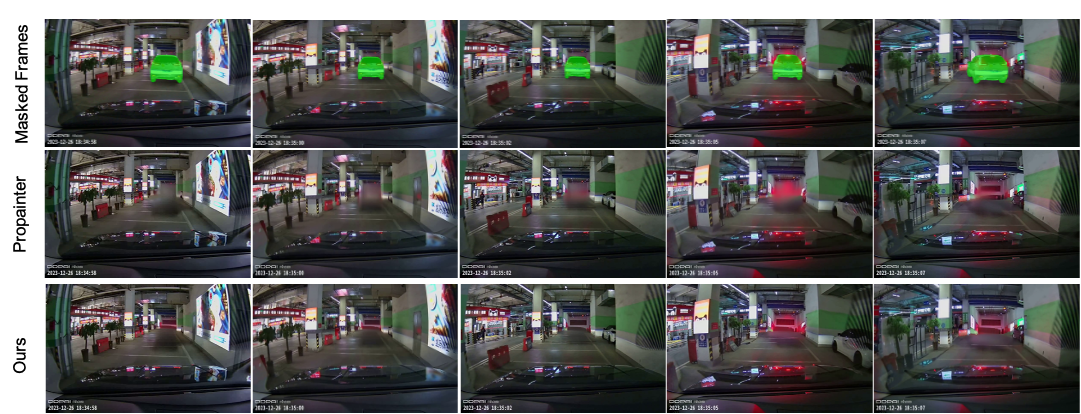}
    \vspace{-8pt}
\end{center}
   \caption{Temporal consistency comparison between \textbf{DiffuEraser} and Propainter.}
   \vspace{-2pt}
\label{fig:14}
\end{figure*}

\textbf{Qualitative Comparison.} Figure \ref{fig:1} illustrates a comparison between our model and Propainter both in texture quality and temporal consistency. For more comparison results, see Figure \ref{fig:11},\ref{fig:10},\ref{fig:13},\ref{fig:14}. Our model effectively propagates known pixels—those that appear in some masked frames—to all frames, while also generating unknown pixels—those that never appear in any masked frames—with high consistency and stability. 



\section{Conclusion and Discussion}

In this paper, we introduce \textbf{DiffuEraser}, a video inpainting model based on stable diffusion. We address the video inpainting task by decomposing it into three sub-problems: propagation of known pixels (pixels appearing in some masked frames), generation of unknown pixels (pixels never appearing in any masked frames), and maintaining temporal consistency of the completed content. For each sub-problem, we propose tailored solutions.

For the generation of unknown pixels, the powerful generative capabilities of the stable diffusion model help DiffuEraser effectively overcome the blurring and mosaic issues prevalent in Transformer-based models. Additionally, we mitigate the inherent hallucinations of stable diffusion models by incorporating priors, ensuring more accurate and realistic inpainting results.

In terms of propagating known pixels, the motion module within the denoising UNet, combined with the enhanced propagation properties provided by priors, ensures the sufficient and consistent propagation of known pixels across frames. This prevents conflicts between the completed content and the unmasked regions, thereby improving the correctness and stability of the results.

To address temporal inconsistencies between clips for long-sequence inference, we expand the temporal receptive field for both prior model and DiffuEraser, significantly enhancing the consistency of completed content across all frames. Furthermore, we leverage the temporal smoothing property of VDM to further enhance temporal coherence at the intersections between clips.

The concepts of incorporating priors and the methods to improve temporal consistency for long-sequence inference are also applicable to a variety of other video editing tasks, such as object replacement and local stylization. These applications will be further explored in future works. Experimental results demonstrate that \textbf{DiffuEraser} outperforms state-of-the-art methods in both content completeness and temporal consistency, establishing it as a superior approach for video inpainting tasks.


{\small
\bibliographystyle{ieee_fullname}
\bibliography{egbib}

\begin{thebibliography}{10}\itemsep=-1pt

\bibitem{brooks2022instructpix2pix}
Tim Brooks, Aleksander Holynski, and Alexei~A Efros.
\newblock Instructpix2pix: Learning to follow image editing instructions.
\newblock {\em arXiv preprint arXiv:2211.09800}, 2022.

\bibitem{chen2024panda70m}
Tsai-Shien Chen, Aliaksandr Siarohin, Willi Menapace, Ekaterina Deyneka, Hsiang-wei Chao, Byung~Eun Jeon, Yuwei Fang, Hsin-Ying Lee, Jian Ren, Ming-Hsuan Yang, and Sergey Tulyakov.
\newblock Panda-70m: Captioning 70m videos with multiple cross-modality teachers.
\newblock In {\em Proceedings of the IEEE/CVF Conference on Computer Vision and Pattern Recognition}, 2024.

\bibitem{chen2023controlavideo}
Weifeng Chen, Jie Wu, Pan Xie, Hefeng Wu, Jiashi Li, Xin Xia, Xuefeng Xiao, and Liang Lin.
\newblock Control-a-video: Controllable text-to-video generation with diffusion models, 2023.

\bibitem{esser2023structurecontentguidedvideosynthesis}
Patrick Esser, Johnathan Chiu, Parmida Atighehchian, Jonathan Granskog, and Anastasis Germanidis.
\newblock Structure and content-guided video synthesis with diffusion models, 2023.

\bibitem{ramesh2022}
Aditya~Ramesh et al.
\newblock Hierarchical text-conditional image generation with clip latents, 2022.

\bibitem{gal2022textual}
Rinon Gal, Yuval Alaluf, Yuval Atzmon, Or Patashnik, Amit~H. Bermano, Gal Chechik, and Daniel Cohen-Or.
\newblock An image is worth one word: Personalizing text-to-image generation using textual inversion, 2022.

\bibitem{Gao-ECCV-FGVC}
Chen Gao, Ayush Saraf, Jia-Bin Huang, and Johannes Kopf.
\newblock Flow-edge guided video completion.
\newblock In {\em Proc. European Conference on Computer Vision (ECCV)}, 2020.

\bibitem{ge2024preservecorrelationnoiseprior}
Songwei Ge, Seungjun Nah, Guilin Liu, Tyler Poon, Andrew Tao, Bryan Catanzaro, David Jacobs, Jia-Bin Huang, Ming-Yu Liu, and Yogesh Balaji.
\newblock Preserve your own correlation: A noise prior for video diffusion models, 2024.

\bibitem{gu2024advanced}
Bohai Gu, Hao Luo, Song Guo, and Peiran Dong.
\newblock Advanced video inpainting using optical flow-guided efficient diffusion.
\newblock {\em arXiv preprint arXiv:2412.00857}, 2024.

\bibitem{reuse2023}
Jiaxi Gu, Shicong Wang, Haoyu Zhao, Tianyi Lu, Xing Zhang, Zuxuan Wu, Songcen Xu, Wei Zhang, Yu-Gang Jiang, and Hang Xu.
\newblock Reuse and diffuse: Iterative denoising for text-to-video generation.

\bibitem{guo2023animatediff}
Yuwei Guo, Ceyuan Yang, Anyi Rao, Zhengyang Liang, Yaohui Wang, Yu Qiao, Maneesh Agrawala, Dahua Lin, and Bo Dai.
\newblock Animatediff: Animate your personalized text-to-image diffusion models without specific tuning.
\newblock {\em International Conference on Learning Representations}, 2024.

\bibitem{hertz2022prompt}
Amir Hertz, Ron Mokady, Jay Tenenbaum, Kfir Aberman, Yael Pritch, and Daniel Cohen-Or.
\newblock Prompt-to-prompt image editing with cross attention control.
\newblock {\em arXiv preprint arXiv:2208.01626}, 2022.

\bibitem{ho2022imagenvideohighdefinition}
Jonathan Ho, William Chan, Chitwan Saharia, Jay Whang, Ruiqi Gao, Alexey Gritsenko, Diederik~P. Kingma, Ben Poole, Mohammad Norouzi, David~J. Fleet, and Tim Salimans.
\newblock Imagen video: High definition video generation with diffusion models, 2022.

\bibitem{ho2020denoising}
Jonathan Ho, Ajay Jain, and Pieter Abbeel.
\newblock Denoising diffusion probabilistic models.
\newblock {\em arXiv preprint arxiv:2006.11239}, 2020.

\bibitem{ho2022video}
Jonathan Ho, Tim Salimans, Alexey Gritsenko, William Chan, Mohammad Norouzi, and David~J Fleet.
\newblock Video diffusion models.
\newblock {\em arXiv:2204.03458}, 2022.

\bibitem{ju2024brushnet}
Xuan Ju, Xian Liu, Xintao Wang, Yuxuan Bian, Ying Shan, and Qiang Xu.
\newblock Brushnet: A plug-and-play image inpainting model with decomposed dual-branch diffusion, 2024.

\bibitem{lee2024videodiffusionmodelsstrong}
Minhyeok Lee, Suhwan Cho, Chajin Shin, Jungho Lee, Sunghun Yang, and Sangyoun Lee.
\newblock Video diffusion models are strong video inpainter, 2024.

\bibitem{liCvpr22vInpainting}
Zhen Li, Cheng-Ze Lu, Jianhua Qin, Chun-Le Guo, and Ming-Ming Cheng.
\newblock Towards an end-to-end framework for flow-guided video inpainting.
\newblock In {\em IEEE Conference on Computer Vision and Pattern Recognition (CVPR)}, 2022.

\bibitem{liew2023magicedit}
Jun~Hao Liew, Hanshu Yan, Jianfeng Zhang, Zhongcong Xu, and Jiashi Feng.
\newblock Magicedit: High-fidelity and temporally coherent video editing.
\newblock In {\em arXiv}, 2023.

\bibitem{Liu_2021_FuseFormer}
Rui Liu, Hanming Deng, Yangyi Huang, Xiaoyu Shi, Lewei Lu, Wenxiu Sun, Xiaogang Wang, Jifeng Dai, and Hongsheng Li.
\newblock Fuseformer: Fusing fine-grained information in transformers for video inpainting.
\newblock In {\em International Conference on Computer Vision (ICCV)}, 2021.

\bibitem{liu2023videop2p}
Shaoteng Liu, Yuechen Zhang, Wenbo Li, Zhe Lin, and Jiaya Jia.
\newblock Video-p2p: Video editing with cross-attention control, 2023.

\bibitem{mokady2022null}
Ron Mokady, Amir Hertz, Kfir Aberman, Yael Pritch, and Daniel Cohen-Or.
\newblock Null-text inversion for editing real images using guided diffusion models.
\newblock {\em arXiv preprint arXiv:2211.09794}, 2022.

\bibitem{molad2023dreamixvideodiffusionmodels}
Eyal Molad, Eliahu Horwitz, Dani Valevski, Alex~Rav Acha, Yossi Matias, Yael Pritch, Yaniv Leviathan, and Yedid Hoshen.
\newblock Dreamix: Video diffusion models are general video editors, 2023.

\bibitem{mou2023t2i}
Chong Mou, Xintao Wang, Liangbin Xie, Yanze Wu, Jian Zhang, Zhongang Qi, Ying Shan, and Xiaohu Qie.
\newblock T2i-adapter: Learning adapters to dig out more controllable ability for text-to-image diffusion models.
\newblock {\em arXiv preprint arXiv:2302.08453}, 2023.

\bibitem{qi2023fatezero}
Chenyang Qi, Xiaodong Cun, Yong Zhang, Chenyang Lei, Xintao Wang, Ying Shan, and Qifeng Chen.
\newblock Fatezero: Fusing attentions for zero-shot text-based video editing.
\newblock {\em arXiv:2303.09535}, 2023.

\bibitem{quan2024deeplearningbasedimagevideo}
Weize Quan, Jiaxi Chen, Yanli Liu, Dong-Ming Yan, and Peter Wonka.
\newblock Deep learning-based image and video inpainting: A survey, 2024.

\bibitem{rombach2021highresolution}
Robin Rombach, Andreas Blattmann, Dominik Lorenz, Patrick Esser, and Björn Ommer.
\newblock High-resolution image synthesis with latent diffusion models, 2021.

\bibitem{ruiz2022dreambooth}
Nataniel Ruiz, Yuanzhen Li, Varun Jampani, Yael Pritch, Michael Rubinstein, and Kfir Aberman.
\newblock Dreambooth: Fine tuning text-to-image diffusion models for subject-driven generation.
\newblock In {\em arXiv preprint arxiv:2208.12242}, 2022.

\bibitem{saharia2022photorealistictexttoimagediffusionmodels}
Chitwan Saharia, William Chan, Saurabh Saxena, Lala Li, Jay Whang, Emily Denton, Seyed Kamyar~Seyed Ghasemipour, Burcu~Karagol Ayan, S.~Sara Mahdavi, Rapha~Gontijo Lopes, Tim Salimans, Jonathan Ho, David~J Fleet, and Mohammad Norouzi.
\newblock Photorealistic text-to-image diffusion models with deep language understanding, 2022.

\bibitem{Shi_2024_CVPR}
Fengyuan Shi, Jiaxi Gu, Hang Xu, Songcen Xu, Wei Zhang, and Limin Wang.
\newblock Bivdiff: A training-free framework for general-purpose video synthesis via bridging image and video diffusion models.
\newblock In {\em Proceedings of the IEEE/CVF Conference on Computer Vision and Pattern Recognition (CVPR)}, pages 7393--7402, June 2024.

\bibitem{singer2022makeavideotexttovideogenerationtextvideo}
Uriel Singer, Adam Polyak, Thomas Hayes, Xi Yin, Jie An, Songyang Zhang, Qiyuan Hu, Harry Yang, Oron Ashual, Oran Gafni, Devi Parikh, Sonal Gupta, and Yaniv Taigman.
\newblock Make-a-video: Text-to-video generation without text-video data, 2022.

\bibitem{sohldickstein2015deepunsupervisedlearningusing}
Jascha Sohl-Dickstein, Eric~A. Weiss, Niru Maheswaranathan, and Surya Ganguli.
\newblock Deep unsupervised learning using nonequilibrium thermodynamics, 2015.

\bibitem{song2020denoising}
Jiaming Song, Chenlin Meng, and Stefano Ermon.
\newblock Denoising diffusion implicit models.
\newblock {\em arXiv:2010.02502}, October 2020.

\bibitem{song2021scorebased}
Yang Song, Jascha Sohl-Dickstein, Diederik~P Kingma, Abhishek Kumar, Stefano Ermon, and Ben Poole.
\newblock Score-based generative modeling through stochastic differential equations.
\newblock In {\em International Conference on Learning Representations}, 2021.

\bibitem{wang2024phased}
Fu-Yun Wang, Zhaoyang Huang, Alexander~William Bergman, Dazhong Shen, Peng Gao, Michael Lingelbach, Keqiang Sun, Weikang Bian, Guanglu Song, Yu Liu, et~al.
\newblock Phased consistency model.
\newblock {\em arXiv preprint arXiv:2405.18407}, 2024.

\bibitem{wang2023videocomposercompositionalvideosynthesis}
Xiang Wang, Hangjie Yuan, Shiwei Zhang, Dayou Chen, Jiuniu Wang, Yingya Zhang, Yujun Shen, Deli Zhao, and Jingren Zhou.
\newblock Videocomposer: Compositional video synthesis with motion controllability, 2023.

\bibitem{wu2024lgvi}
Jianzong Wu, Xiangtai Li, Chenyang Si, Shangchen Zhou, Jingkang Yang, Jiangning Zhang, Yining Li, Kai Chen, Yunhai Tong, Ziwei Liu, et~al.
\newblock Towards language-driven video inpainting via multimodal large language models.
\newblock {\em arXiv preprint arXiv:2401.10226}, 2024.

\bibitem{wu2023tune}
Jay~Zhangjie Wu, Yixiao Ge, Xintao Wang, Stan~Weixian Lei, Yuchao Gu, Yufei Shi, Wynne Hsu, Ying Shan, Xiaohu Qie, and Mike~Zheng Shou.
\newblock Tune-a-video: One-shot tuning of image diffusion models for text-to-video generation.
\newblock In {\em Proceedings of the IEEE/CVF International Conference on Computer Vision}, pages 7623--7633, 2023.

\bibitem{xing2023make}
Jinbo Xing, Menghan Xia, Yuxin Liu, Yuechen Zhang, Yong Zhang, Yingqing He, Hanyuan Liu, Haoxin Chen, Xiaodong Cun, Xintao Wang, et~al.
\newblock Make-your-video: Customized video generation using textual and structural guidance.
\newblock {\em arXiv preprint arXiv:2306.00943}, 2023.

\bibitem{yan2020sttn}
Yanhong Zeng, Jianlong Fu, and Hongyang Chao.
\newblock Learning joint spatial-temporal transformations for video inpainting.
\newblock In {\em The Proceedings of the European Conference on Computer Vision (ECCV)}, 2020.

\bibitem{zhang2022flow}
Kaidong Zhang, Jingjing Fu, and Dong Liu.
\newblock Flow-guided transformer for video inpainting.
\newblock In {\em European Conference on Computer Vision}, pages 74--90. Springer, 2022.

\bibitem{Zhang_2022_CVPR}
Kaidong Zhang, Jingjing Fu, and Dong Liu.
\newblock Inertia-guided flow completion and style fusion for video inpainting.
\newblock In {\em Proceedings of the IEEE/CVF Conference on Computer Vision and Pattern Recognition (CVPR)}, pages 5982--5991, June 2022.

\bibitem{zhang2023adding}
Lvmin Zhang, Anyi Rao, and Maneesh Agrawala.
\newblock Adding conditional control to text-to-image diffusion models, 2023.

\bibitem{zhang2023controlvideo}
Yabo Zhang, Yuxiang Wei, Dongsheng Jiang, Xiaopeng Zhang, Wangmeng Zuo, and Qi Tian.
\newblock Controlvideo: Training-free controllable text-to-video generation.
\newblock {\em arXiv preprint arXiv:2305.13077}, 2023.

\bibitem{zhang2023avid}
Zhixing Zhang, Bichen Wu, Xiaoyan Wang, Yaqiao Luo, Luxin Zhang, Yinan Zhao, Peter Vajda, Dimitris Metaxas, and Licheng Yu.
\newblock Avid: Any-length video inpainting with diffusion model.
\newblock {\em arXiv preprint arXiv:2312.03816}, 2023.

\bibitem{zhou2023propainter}
Shangchen Zhou, Chongyi Li, Kelvin~C.K Chan, and Chen~Change Loy.
\newblock {ProPainter}: Improving propagation and transformer for video inpainting.
\newblock In {\em Proceedings of IEEE International Conference on Computer Vision (ICCV)}, 2023.

\bibitem{Zi2024CoCoCo}
Bojia Zi, Shihao Zhao, Xianbiao Qi, Jianan Wang, Yukai Shi, Qianyu Chen, Bin Liang, Kam-Fai Wong, and Lei Zhang.
\newblock Cococo: Improving text-guided video inpainting for better consistency, controllability and compatibility.
\newblock {\em ArXiv}, abs/2403.12035, 2024.

\end{thebibliography}
}

\end{document}